\documentclass[conference]{IEEEtran}
\IEEEoverridecommandlockouts
\usepackage{cite}
\usepackage{amsmath,amssymb,amsfonts}
\usepackage{booktabs}
\usepackage{algorithmic}
\usepackage{graphicx}
\usepackage{textcomp}
\usepackage{xcolor}
\usepackage{placeins}
\usepackage{multirow}
\usepackage{fancyhdr}
\def\BibTeX{{\rm B\kern-.05em{\sc i\kern-.025em b}\kern-.08em
    T\kern-.1667em\lower.7ex\hbox{E}\kern-.125emX}}
\usepackage{fontspec}
\newfontfamily\bengalifont{NotoSansBengali-Regular.ttf}[Path=./, Script=Bengali]
\newcommand{\bn}[1]{{\bengalifont #1}}
\usepackage{titlesec}
\titlespacing*{\section}{0pt}{4pt plus 1pt minus 1pt}{2pt plus 1pt minus 1pt}
\titlespacing*{\subsection}{0pt}{3pt plus 1pt minus 1pt}{1pt plus 1pt minus 1pt}

\IEEEoverridecommandlockouts

\newcommand{\ieeecopyrighttext}{%
  \textcopyright{} 2026 IEEE. Personal use of this material is permitted. Permission from IEEE must be obtained for all other uses, in any current or future media, including reprinting/republishing this material for advertising or promotional purposes, creating new collective works, for resale or redistribution to servers or lists, or reuse of any copyrighted component of this work in other works.}
\begin{document}

\title{BanglaRhet: Benchmarking Classical and Transformer Models for Rhetorical and Persuasion Detection in Bangla Political Speech}

\author{\IEEEauthorblockN{1\textsuperscript{st} Rohit Kumar Sen}
\IEEEauthorblockA{\textit{Department of Computer Science and Engineering} \\
\textit{North East University Bangladesh}\\
Sylhet, Bangladesh \\
rohit.k.sen.neub@gmail.com}
\and
\IEEEauthorblockN{2\textsuperscript{nd} Anik Chowdhury}
\IEEEauthorblockA{\textit{Department of Computer Science and Engineering} \\
\textit{North East University Bangladesh}\\
Sylhet, Bangladesh \\
anikchowdhuryraj.neub@gmail.com}}

\maketitle

\fancyhf{}  

\fancyhead[L]{\fontsize{9}{10}\selectfont 2026 2nd International Conference on Advances in Computing, Communication, Electrical, and Smart Systems (iCACCESS)\\ 02--03 October 2026, Dhaka, Bangladesh}

\fancyfoot[L]{%
  \parbox[t]{\textwidth}{%
    \fontsize{7}{8}\selectfont \ieeecopyrighttext\par
    \vspace{2pt}
    \fontsize{9}{10}\selectfont 979-8-3195-1375-5/26/\$31.00~\copyright~2026 IEEE}}

\renewcommand{\headrulewidth}{0pt}
\renewcommand{\footrulewidth}{0pt}

\thispagestyle{fancy}  

\begin{abstract}
Political discourse often uses rhetorical and persuasive language to frame narratives, influence public opinion, and mobilize audiences. While Bangla natural language processing has made progress in sentiment analysis, opinion mining, and pretrained language modeling, systematic benchmarking of transformer models for fine-grained rhetorical and persuasion technique detection in Bangla political speech remains largely underexplored. This paper presents a benchmark study of transformer-based models for detecting rhetorical form and persuasive intent in Bangla political discourse. Using BanglaRhet, a manually annotated corpus of 30,289 Bangla political speech segments collected from publicly available political news sources and speech-related articles, we formulate two supervised single-label classification tasks: rhetorical technique detection and persuasion technique detection. The rhetorical task captures stylistic forms such as contrast, repetition, exaggeration, metaphor, and rhetorical questions, while the persuasion task captures communicative functions such as blame assignment, call to action, unity call, moral appeal, emotional appeal, and logical appeal. We evaluate four transformer-based models, BanglaBERT, Bangla-BERT-Base, SahajBERT, and XLM-RoBERTa-Base, against classical TF-IDF baselines. BanglaBERT achieves the highest performance, with 65.40\% macro-F1 for rhetorical technique detection and 66.46\% macro-F1 for persuasion technique detection, outperforming the best tuned classical baseline by 19.2 macro-F1 points for rhetorical technique detection and 13.8 macro-F1 points for persuasion technique detection. Class-level analysis indicates that model errors are mainly associated with semantic overlap among labels, figurative language, and class imbalance. The results provide initial benchmark baselines for Bangla rhetorical and persuasion-aware political discourse analysis and highlight the need for future work on context-aware and multi-label modeling.
\end{abstract}


\section{Introduction}

Political discourse, including speeches, debates, and news commentary, is a powerful medium for shaping public opinion, framing ideological positions, and influencing social behavior \cite{sultonov2025use, salih2024persuasive}. Political texts often rely on rhetorical and persuasive strategies to strengthen arguments, assign responsibility, appeal to values, and mobilize audiences \cite{afzal2021using, sultonov2025use}. Automated analysis of these strategies can support research in political science, the social sciences, computational linguistics, and natural language processing (NLP).

In this study, we distinguish between two complementary aspects of political language: rhetorical techniques, which describe how a statement is expressed, and persuasion techniques, which describe the communicative purpose of the statement. Detecting these aspects is challenging because political language often contains figurative expressions, emotionally charged wording, implicit intent, and overlapping categories. As a result, models must capture stylistic and pragmatic cues beyond surface-level meaning.

Despite recent progress in Bangla NLP, existing Bangla political text classification studies have mainly focused on sentiment analysis, opinion mining, polarity detection, and related tasks \cite{arman2025evaluating, faria2024motamot}. Although useful, these tasks primarily capture broad attitudes or opinions and do not directly model the rhetorical structure or persuasive function of political statements. Fine-grained, technique-level taxonomies exist for related English tasks such as propaganda technique detection \cite{dasanmartino2020semeval}, but no equivalent resource exists for Bangla. Consequently, benchmark evaluation for fine-grained rhetorical and persuasion-aware analysis in Bangla political discourse remains largely underexplored.

To address this gap, this paper presents a transformer-based benchmark study using \textbf{BanglaRhet}, a manually annotated corpus of 30,289 Bangla political speech segments collected from publicly available political news sources and speech-related articles \cite{prothomalo, faria2024motamot}. We evaluate Bangla-specific transformer models, including BanglaBERT \cite{csebuetnlp}, SahajBERT \cite{sahajbert}, and Bangla-BERT-Base \cite{banglabertbase}, along with the multilingual baseline XLM-RoBERTa-Base \cite{xlmr}.

The main contributions of this work are: (i) a benchmark framework for fine-grained rhetorical and persuasion technique detection tasks in Bangla political discourse; (ii) an evaluation of Bangla-specific and multilingual transformer models using 30,289 manually annotated political speech segments; (iii) an empirical comparison against classical TF-IDF-based baselines confirming the necessity of transformer-based representations for this task; and (iv) class-level analysis of challenges caused by overlapping label semantics, figurative usage, and class imbalance.

\section{Background and Related Work}
\label{sec:background}

\subsection{Rhetorical and Persuasive Language in Political Discourse}

Political communication often relies on rhetorical and persuasive strategies to influence audience perception, frame ideological positions, and strengthen arguments. Rhetorical techniques describe how political language is expressed, while persuasion techniques describe the communicative intent or strategic function of a statement \cite{sultonov2025use, salih2024persuasive}. Based on this distinction, this study models rhetorical categories such as contrast, repetition, metaphor, exaggeration, and rhetorical questions, and persuasion categories such as blame assignment, moral appeal, logical appeal, emotional appeal, and call to action.

\subsection{Political Text Analysis and Persuasion Detection}

Computational political text analysis has been studied through sentiment analysis, ideology detection, stance detection, argument mining, and persuasion detection. Prior work has analyzed political sentiment and public opinion from news texts, social media posts, and online comments \cite{hossain2024political}, but such polarity scores cannot distinguish, for example, a metaphor-driven appeal from a direct call to action expressed with the same sentiment. Transformer-based models have also been applied to political communication tasks, such as PoliBERT for classifying political social media messages \cite{gupta2020polibert}, though this targets short social-media posts using coarse functional categories rather than technique-level rhetorical form.

Persuasion-related studies include the CORPS corpus, which used political speeches and audience applause as a weak, effect-level proxy for persuasiveness rather than a technique-level label \cite{guerini2008corps}, and traditional persuasiveness prediction methods that output a single scalar score rather than a categorical taxonomy \cite{strapparava2010predicting}. Recent persuasion detection work in advertising, online campaigns, and social media adopts finer-grained technique labels \cite{meguellati2025persuasion}, but remains limited to English, non-political domains. Closest to our setting is SemEval-2020 Task 11, which introduced a 14-category taxonomy of propaganda techniques (e.g., loaded language, repetition, exaggeration, appeal to fear) for English news, with transformer-based models performing best \cite{dasanmartino2020semeval}. This confirms that fine-grained technique taxonomies are learnable, but no equivalent exists for Bangla, and none of the above jointly model stylistic form and communicative function as we do.

\subsection{Bangla Political NLP and Existing Resources}

Bangla political NLP research has mainly focused on sentiment analysis, opinion mining, polarity detection, ideology, and stance-related tasks. The \textit{Motamot} dataset provides Bangla political news texts annotated for binary sentiment and supports comparative evaluation of Bangla NLP models \cite{faria2024motamot}; sentiment, however, is a downstream byproduct of rhetorical strategy, so two segments can share sentiment while using entirely different techniques. Other studies analyze political sentiment from YouTube and news comments \cite{arman2025evaluating}, targeting audience reaction rather than the speaker's own discourse, and political ideology or stance prediction has been explored with classical machine learning methods that model leaning rather than rhetorical device \cite{tasnim2021political}. Bengali transnational political discourse resources such as BTPD \cite{das2025btpd} hand-curate discussion from online Q\&A and social platforms and characterize it through unsupervised topic modeling; the resource is not annotated for rhetorical or persuasive technique and targets informal community discussion rather than formal political speech.

Overall, prior Bangla political NLP operates almost exclusively at the sentiment or stance level, while technique-level rhetorical and persuasion analysis, established for English propaganda detection \cite{dasanmartino2020semeval}, has no Bangla counterpart. BanglaRhet addresses this by introducing (i) a Bangla-specific taxonomy spanning nine rhetorical and seven persuasion techniques, (ii) a manually annotated benchmark of 30,289 political speech segments, to our knowledge the first of its kind for Bangla, and (iii) an evaluation of whether Bangla-specific pretraining offers a measurable advantage over multilingual pretraining on this fine-grained task.

\section{Benchmark Dataset and Task Formulation}
\label{sec:dataset}

\subsection{Data Collection and Segment Extraction}

BanglaRhet was built from publicly available Bangla political news and speech-related articles. We used 1,890 links from the existing \textit{Motamot} dataset \cite{faria2024motamot}, supplemented with links from \textit{Prothom Alo} \cite{prothomalo} and \textit{Manab Zamin} \cite{mzamin} covering 2014--2025. In total, 15,032 articles were successfully processed into usable segments after removing inaccessible, duplicate, malformed, or non-political pages.

Articles were manually reviewed to identify political speech segments: a segment was included if it contained a direct quotation or reported statement attributed to a political figure or institution, by name or by role or position (e.g., a local MP), as characterized by the source outlet; where the speaker's identity could not be determined, the segment was retained and marked \texttt{unknown\_speaker}. Journalistic analysis, background context, and commentary from non-political actors were excluded, except where retained to construct the \texttt{None} class. Extraction was performed collaboratively, with segments periodically cross-checked and ambiguous cases resolved through discussion. Passages exceeding the 128-token input limit were condensed, with both authors manually verifying that rhetorical and persuasive content was preserved.

\subsection{Task Formulation, Labels, and Annotation}

We formulate two independent classification tasks per segment $x$: rhetorical technique $y_r$ and persuasion technique $y_p$, each assigned its dominant label; multi-label modeling is left for future work. The rhetorical task has nine labels (\textit{Contrast, Emotional Language, Repetition, Exaggeration, Anecdote, Rhetorical Question, Metaphor, Parallelism}, and \texttt{None}); the persuasion task has seven (\textit{Blame Assignment, Call to Action, Unity Call, Emotional Appeal, Moral Appeal, Logical Appeal}, and \texttt{None}). Table~\ref{tab:label-defs} gives the operational definition of each label. When a segment plausibly exhibited more than one technique, annotators assigned the single \emph{dominant} label, i.e., the one most central to the segment's overall effect, with disagreements resolved by discussion.

\begin{table}[t]
\centering
\caption{Operational label definitions.}
\label{tab:label-defs}
\small
\begin{tabular}{p{2.1cm}p{5.5cm}}
\toprule
\multicolumn{2}{l}{\textbf{Rhetorical Techniques}} \\
\midrule
Contrast & Opposing ideas presented to emphasize difference. \\
Emotional Lang. & Emotionally charged wording to influence feeling. \\
Repetition & Repeated words/phrases to reinforce a message. \\
Exaggeration & Deliberate overstatement for rhetorical impact. \\
Anecdote & Concrete example or personal experience as support. \\
Rhet. Question & Question posed to provoke reflection, not an answer. \\
Metaphor & Figurative comparison for imagery/interpretation. \\
Parallelism & Repeated grammatical structure for emphasis. \\
None & No predefined rhetorical technique present. \\
\midrule
\multicolumn{2}{l}{\textbf{Persuasion Techniques}} \\
\midrule
Blame Assignment & Attributing fault to an opposing party. \\
Call to Action & Urging the audience toward a specific action. \\
Unity Call & Promoting solidarity or collective identity. \\
Emotional Appeal & Persuasion via emotional reaction over logic. \\
Moral Appeal & Appeal to ethics, justice, or moral duty. \\
Logical Appeal & Persuasion via facts, evidence, or reasoning. \\
None & No predefined persuasion technique present. \\
\bottomrule
\end{tabular}
\end{table}

A total of 30,289 segments were manually annotated by two authors with prior study of political discourse. Both jointly annotated a pilot sample to converge on shared label boundaries, particularly for conceptually related pairs (e.g., Emotional Language vs.\ Emotional Appeal), continuing until disagreement stabilized before annotating the full corpus. A segment was labeled \texttt{None} when it showed no discernible instance of any technique, or, less commonly, when it contained borderline language not clearly satisfying any single label's definition. Inter-annotator reliability, measured on a random 500-segment subset before consensus resolution, showed moderate-to-substantial agreement: Cohen's $\kappa = 0.67$ for rhetorical technique and $0.71$ for persuasion technique.

\subsection{Benchmark Statistics and Imbalance}
\label{sec:dataset_analysis}
\begin{figure}[ht]
\centering
\begin{minipage}{0.49\linewidth}
\centering
\includegraphics[width=\linewidth]{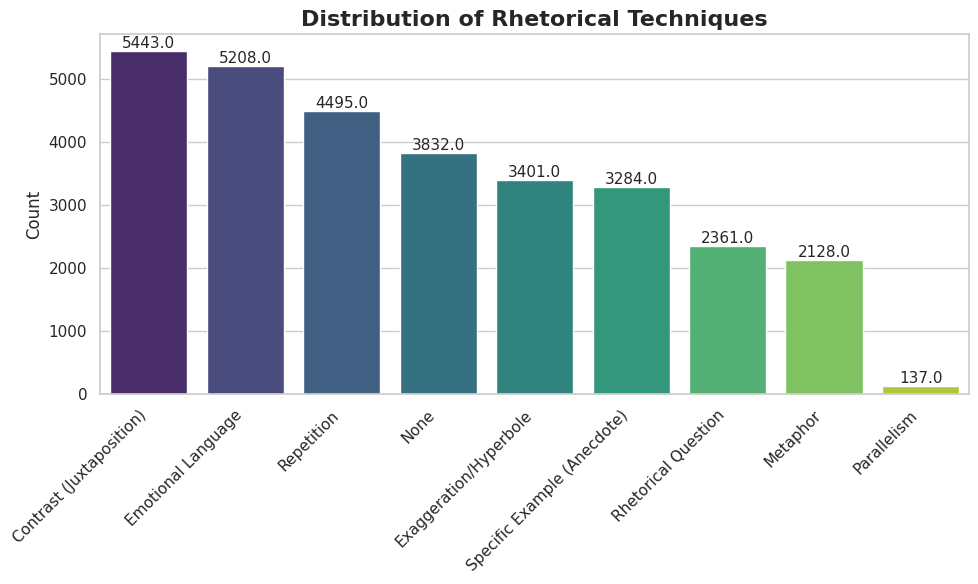}
\small (a) Rhetorical techniques
\end{minipage}
\hfill
\begin{minipage}{0.49\linewidth}
\centering
\includegraphics[width=\linewidth]{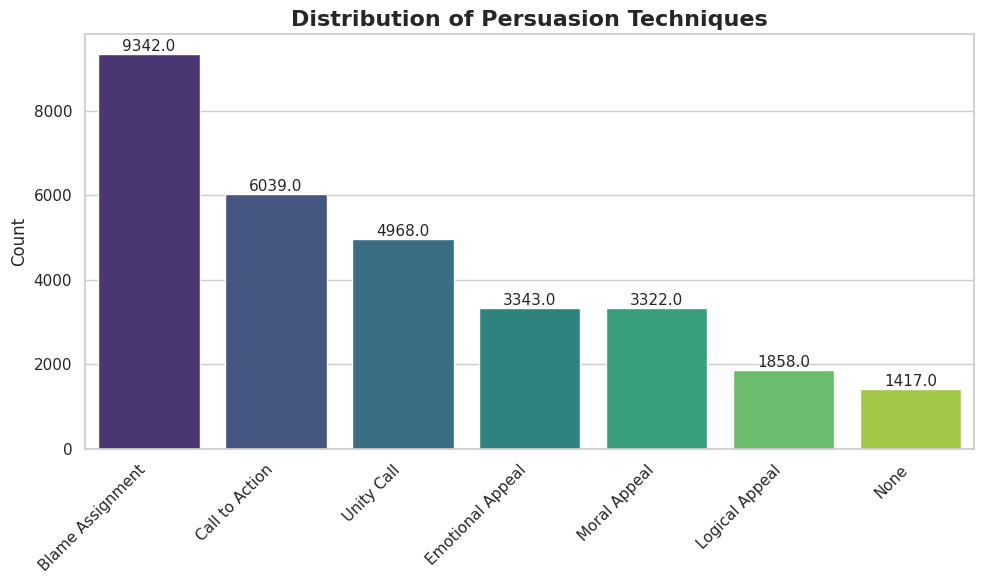}
\small (b) Persuasion techniques
\end{minipage}
\caption{Label distributions in BanglaRhet: (a) rhetorical techniques, (b) persuasion techniques.}
\label{fig:label_dist}
\end{figure}

BanglaRhet contains 30,289 segments drawn from 15,032 sources, averaging 48.34 words per segment, with 9 rhetorical and 7 persuasion labels (each including \texttt{None}). Figure~\ref{fig:label_dist} shows both label distributions. The benchmark is naturally imbalanced: \textit{Contrast}, \textit{Emotional Language}, and \textit{Repetition} dominate the rhetorical dimension while \textit{Parallelism} is rare; \textit{Blame Assignment}, \textit{Call to Action}, and \textit{Unity Call} dominate the persuasion dimension. This imbalance motivates the macro-F1 metric and the class-balancing strategy described in Section~\ref{sec:experiments}.
The rhetorical distribution also shows a sharp discontinuity: consecutive classes differ by at most 1.4x, but \textit{Parallelism} (137 instances) is over 15 times rarer than the next smallest class, \textit{Metaphor} (2{,}128), yielding only about 14 test instances under our split, too few for stable macro-F1; we therefore excluded it rather than oversampling (risk of near duplicate memorization) or merging it with \textit{Repetition}, a structurally distinct category in our taxonomy (Table~\ref{tab:label-defs}).

\section{Experimental Setup}
\label{sec:experiments}

The objective of the experiments is to establish benchmark results for rhetorical and persuasion detection tasks in Bangla political speech. Rather than proposing a new model architecture, this work evaluates whether widely used Bangla-specific and multilingual transformer models can learn fine-grained rhetorical and persuasive patterns under a fixed experimental setting.

\subsection{Task Setup}

The processed benchmark was split 80\%–10\%–10\% into training, validation, and test sets, stratified on the rhetorical labels; persuasion labels were encoded separately over the same splits. Segments were tokenized with each model's tokenizer, padded or truncated to 128 tokens.

\subsection{Class Balancing Strategy}

The original benchmark is naturally imbalanced, with some rhetorical and persuasion techniques occurring much more frequently than others. Critically, this imbalance is not merely per-class but joint: \textit{Blame Assignment}, \textit{Call to Action}, and \textit{Unity Call} together account for 67.2\% of all segments, and they also dominate the co-occurrence counts of the most frequent rhetorical labels (e.g., \textit{Contrast} co-occurs with \textit{Blame Assignment} 2{,}923 times, versus at most 824 times with any other persuasion label). Since uniform per-class downsampling cannot correct this pairwise skew, we instead manually inspected the full rhetorical--persuasion co-occurrence table and reduced the ten most overrepresented pairs, as shown in Table~\ref{tab:downsampling_rules}. \textit{Parallelism} was excluded separately from the experimental benchmark for the reasons discussed in Section~\ref{sec:dataset_analysis}.

\begin{table}[htbp]
\centering
\caption{Controlled downsampling rules applied to overrepresented co-occurrence pairs across the rhetorical and persuasion tasks.}
\label{tab:downsampling_rules}
\small
\setlength{\tabcolsep}{3pt}
\begin{tabular}{|l|l|r|}
\hline
\textbf{Rhetorical Label} & \textbf{Persuasion Label} & \textbf{Removed} \\
\hline
Contrast & Blame Assignment & 2000 \\
Emotional Language & Blame Assignment & 1571 \\
Emotional Language & Call to Action & 447 \\
Exaggeration & Blame Assignment & 537 \\
Repetition & Blame Assignment & 375 \\
Repetition & Call to Action & 1100 \\
None & Call to Action & 300 \\
None & Blame Assignment & 685 \\
None & Unity Call & 300 \\
Anecdote & Unity Call & 272 \\
\hline
\end{tabular}
\end{table}

These reductions were applied through iterative manual inspection of the co-occurrence table rather than a fixed formula, which we acknowledge as a limitation in Section~\ref{sec:limitations}; however, the ablation in Table~\ref{tab:downsampling_ablation} confirms this step was not merely cosmetic, as training without these reductions yields consistently lower macro-F1 across all four evaluated models.

After removing \textit{Parallelism} and applying controlled downsampling, the processed experimental benchmark contained 22,565 samples, while the full annotated BanglaRhet corpus contained 30,289 segments. This processed benchmark was used for the train--validation--test split. During training, class-weighted cross-entropy loss was applied for both tasks to further reduce majority-class bias.

\subsection{Baseline Models}
We evaluate four pretrained transformer models commonly used in Bangla NLP research: \textbf{Bangla-BERT-Base} \cite{banglabertbase}, \textbf{SahajBERT} \cite{sahajbert}, \textbf{BanglaBERT} \cite{csebuetnlp}, and \textbf{XLM-RoBERTa-Base} \cite{xlmr}. The first three are Bangla-specific pretrained models, while XLM-RoBERTa-Base is a multilingual cross-lingual baseline, allowing us to examine whether Bangla-specific pretraining offers advantages over multilingual pretraining for this task; all models were fine-tuned using the standard sequence classification head from the Hugging Face Transformers library. To justify this choice empirically, we also trained two classical baselines, TF-IDF with Logistic Regression and TF-IDF with a linear SVM, with hyperparameters (regularization strength, vocabulary size, n-gram range) selected via grid search on the validation set; as shown in Table~\ref{tab:results}, both are outperformed by every transformer model even after tuning, confirming that this task needs representations beyond bag-of-words features.

\subsection{Training Configuration and Evaluation}

All models were trained for ten epochs using the AdamW optimizer with a learning rate of $2 \times 10^{-5}$, a batch size of 16, and a maximum sequence length of 128 tokens. A single random seed (42) was fixed throughout for data splitting, controlled downsampling, and model training, ensuring a consistent, reproducible comparison across models. The best-performing checkpoint was selected based on validation macro-F1, with class-weighted cross-entropy loss (weights inversely proportional to class frequency) applied for both tasks, which can raise macro-F1 relative to accuracy by improving minority-class performance. Other settings followed the default Hugging Face Trainer configuration; extensive hyperparameter tuning was not performed, as the goal is reproducible benchmark baselines rather than maximizing architecture-specific performance. Models are evaluated on the held-out test set using accuracy and macro-F1, with macro-F1 emphasized given the benchmark's imbalance. The dataset, annotation guidelines, preprocessing scripts, and experimental code will be released in a public repository upon acceptance.

\section{Results and Analysis}
\label{sec:results}

\subsection{Benchmark Results}
Table \ref{tab:results} reports test-set performance for all
evaluated models, including the tuned TF-IDF baselines. Both
classical baselines are outperformed by every transformer model on
both tasks. BanglaBERT achieves the highest performance overall, with 65.40\%
macro-F1 for rhetorical detection and 66.46\% macro-F1 for
persuasion detection, a gain of 19.2 and 13.8 points respectively
over the best tuned classical baseline (Table~\ref{tab:results}). These results suggest that
BanglaBERT's Bangla-specific pretraining is effective for capturing
stylistic and pragmatic cues in Bangla political discourse, although
the moderate macro-F1 scores indicate that both tasks remain
challenging.

\begin{table}[b]
\centering
\caption{Benchmark performance on the test set.}
\label{tab:results}
\begin{tabular}{lcccc}
\hline
\multirow{2}{*}{\textbf{Model}} & \multicolumn{2}{c}{\textbf{Rhetorical}} & \multicolumn{2}{c}{\textbf{Persuasion}} \\
\cline{2-5}
 & \textbf{Acc} & \textbf{Macro-F1} & \textbf{Acc} & \textbf{Macro-F1} \\
\hline
TF-IDF + LR      & 0.4599 & 0.4587 & 0.5321 & 0.5270 \\
TF-IDF + SVM     & 0.4692 & 0.4624 & 0.5241 & 0.5114 \\
Bangla-BERT-Base & 0.5237 & 0.5285 & 0.5733 & 0.5720 \\
SahajBERT        & 0.5853 & 0.5974 & 0.6185 & 0.6227 \\
BanglaBERT       & 0.6460 & 0.6540 & 0.6619 & 0.6646 \\
XLM-RoBERTa-Base & 0.6043 & 0.6110 & 0.6349 & 0.6405 \\
\hline
\end{tabular}
\end{table}

To examine the effect of controlled downsampling, we conducted an ablation experiment without the downsampling rules while keeping the same label set, random seed, class-weighted loss, and training configuration. The rare \textit{Parallelism} class was excluded in both settings to ensure a consistent label space. As shown in Table~\ref{tab:downsampling_ablation}, controlled downsampling improves macro-F1 across all models for both tasks, with the largest single gain observed for BanglaBERT on rhetorical detection.

\begin{table}[t]
\centering
\caption{Effect of controlled downsampling on macro-F1.}
\label{tab:downsampling_ablation}
\small
\setlength{\tabcolsep}{3pt}
\begin{tabular}{|l|cc|cc|}
\hline
\textbf{Model} & \multicolumn{2}{c|}{\textbf{Rhetorical}} & \multicolumn{2}{c|}{\textbf{Persuasion}} \\
 & No DS & DS & No DS & DS \\
\hline
Bangla-BERT-Base & 0.5238 & \textbf{0.5285} & 0.5500 & \textbf{0.5720} \\
SahajBERT & 0.5821 & \textbf{0.5974} & 0.6086 & \textbf{0.6227} \\
BanglaBERT & 0.6214 & \textbf{0.6540} & 0.6523 & \textbf{0.6646} \\
XLM-RoBERTa-Base & 0.6005 & \textbf{0.6110} & 0.6259 & \textbf{0.6405} \\
\hline
\end{tabular}
\end{table}

\subsection{Effect of Source-Level Splitting}
Since segment-level splitting can allow segments from the same source article to appear in both training and test sets, we additionally evaluated BanglaBERT under a source-grouped split, using \texttt{GroupShuffleSplit} on the article source to guarantee no article contributes segments to more than one split. Under the original segment-level split, 61.23\% of test segments and 59.35\% of validation segments shared a source article with the training set. Table~\ref{tab:source_split} compares BanglaBERT's performance under both splits.

\begin{table}[htbp]
\centering
\caption{BanglaBERT macro-F1 under segment-level vs.\ source-level splitting.}
\label{tab:source_split}
\small
\begin{tabular}{|l|c|c|}
\hline
\textbf{Split} & \textbf{Rhetorical} & \textbf{Persuasion} \\
\hline
Segment-level (original) & 0.6540 & 0.6646 \\
Source-level (grouped) & 0.6438 & 0.6538 \\
\hline
\end{tabular}
\end{table}

Despite substantial source overlap under the original split, macro-F1 drops by only 1.02 and 1.08 points for rhetorical and persuasion detection respectively under the leakage-free source-level split. This modest degradation suggests the model is not substantially reliant on memorized source-specific phrasing or speaker style, and that the benchmark results in Table~\ref{tab:results} are not meaningfully inflated by segment-level splitting.

\subsection{Confusion Matrices and Per-Class Analysis}
\begin{figure}[htbp]
\centering
\begin{minipage}{0.49\linewidth}
\centering
\includegraphics[width=\linewidth]{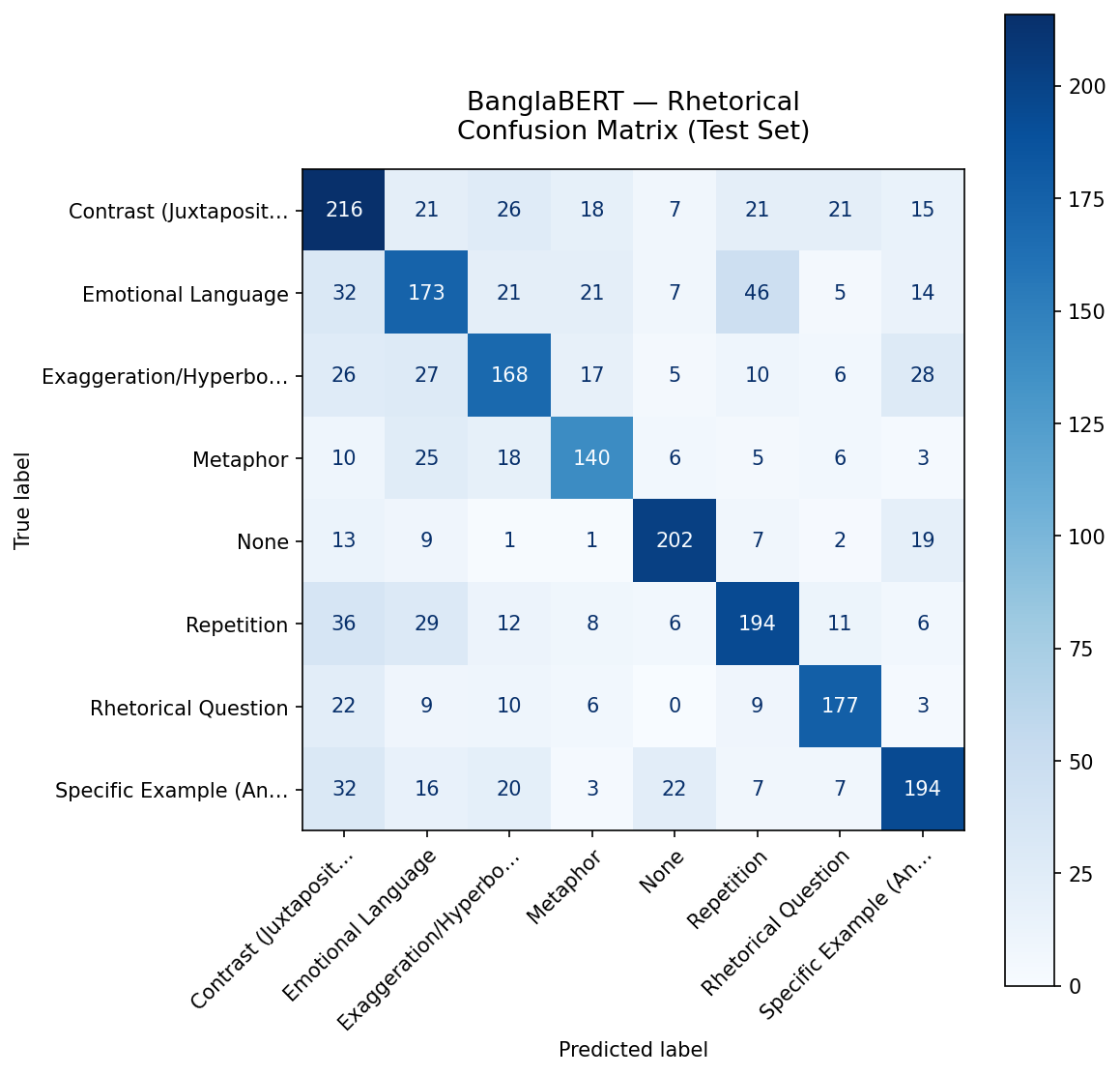}
\small (a) Rhetorical
\end{minipage}
\hfill
\begin{minipage}{0.49\linewidth}
\centering
\includegraphics[width=\linewidth]{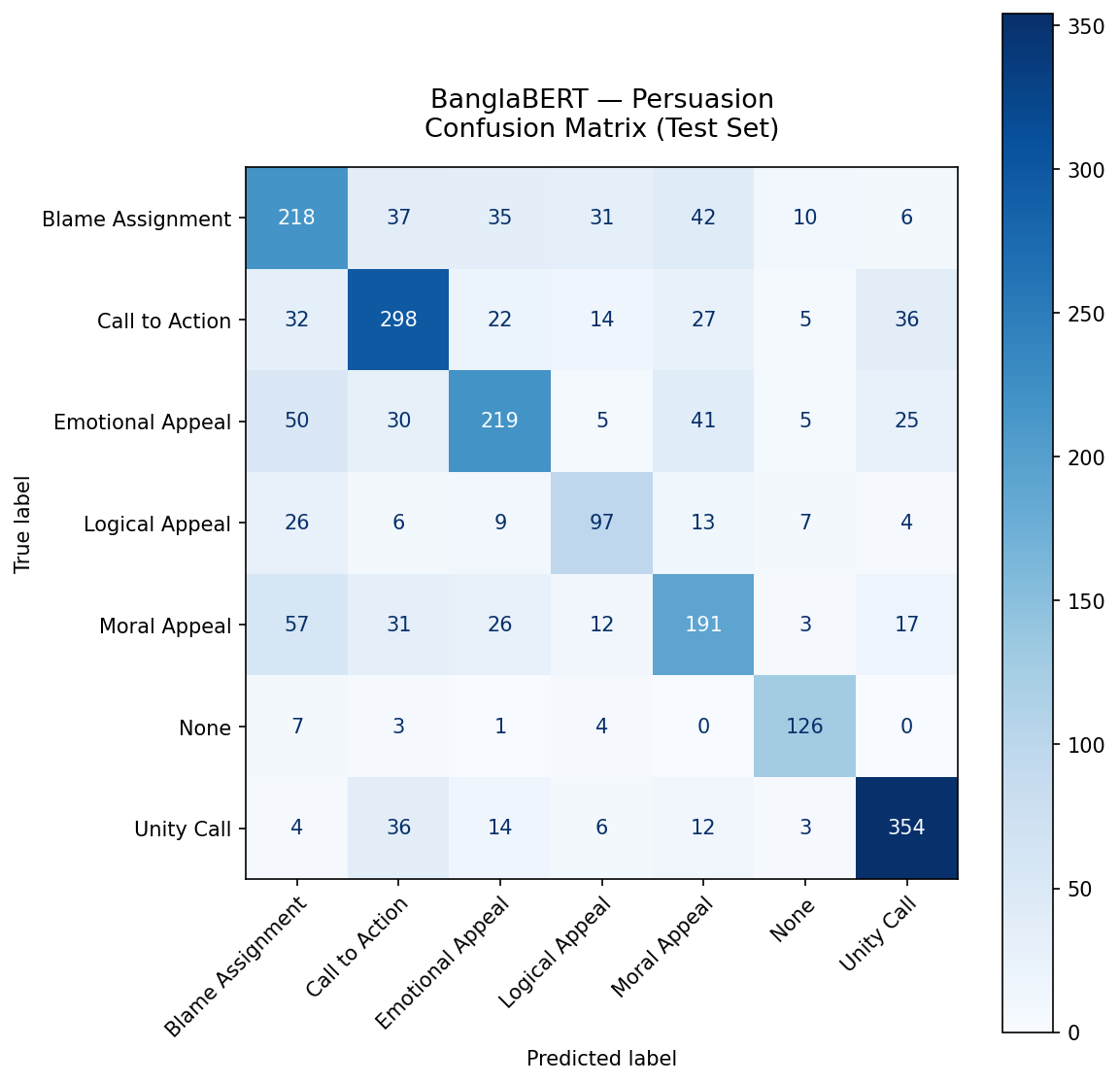}
\small (b) Persuasion
\end{minipage}
\caption{Confusion matrices of BanglaBERT: (a) rhetorical technique detection, (b) persuasion technique detection.}
\label{fig:confusion_matrices}
\end{figure}

To examine class-level behavior, we analyze the confusion matrices of the best-performing model, BanglaBERT. Figure~\ref{fig:confusion_matrices} shows the confusion matrices for rhetorical and persuasion detection.

For rhetorical detection, many errors occur among stylistically related categories. One notable confusion appears between \textit{Emotional Language} and \textit{Repetition}, where 46 \textit{Emotional Language} instances are predicted as \textit{Repetition}. For example, a test segment, \bn{শহীদদের স্বপ্ন এখনো পূরণ হয়নি... আমাদের একত্রিত হয়ে কাজ করতে হবে } (``The martyrs' dream has still not been fulfilled... we must come together and work''), was labeled \textit{Emotional Language} but predicted \textit{Repetition}: the model latches onto the lexical repetition of unity-related phrases, even though annotators judged the segment's dominant effect to be emotional appeal to sacrifice rather than structural reinforcement. Additional confusion occurs between \textit{Repetition} and \textit{Contrast}, and between \textit{Exaggeration} and \textit{Emotional Language}. These patterns indicate that rhetorical devices in political speech often overlap in surface form and expressive function.

For persuasion detection, \textit{Unity Call} and \textit{None} are classified more reliably, while \textit{Blame Assignment}, \textit{Moral Appeal}, and \textit{Emotional Appeal} show more confusion; \textit{Moral Appeal} and \textit{Emotional Appeal} are frequently predicted as \textit{Blame Assignment}, since the explicit naming of a rival party or a specific accusation gives the model a strong blame-oriented surface cue that overrides the underlying moral or emotional framing.

Table~\ref{tab:per_class_banglabert} reports per-class F1 scores computed from the BanglaBERT confusion matrices. The results show stronger performance for categories with clearer lexical or functional cues, such as \textit{None}, \textit{Rhetorical Question}, and \textit{Unity Call}. In contrast, lower F1 scores are observed for semantically overlapping or more difficult categories such as \textit{Emotional Language}, \textit{Blame Assignment}, \textit{Moral Appeal}, and \textit{Logical Appeal}. This suggests that model performance is affected not only by class frequency but also by conceptual overlap among fine-grained political language categories.

\begin{table}[htbp]
\centering
\caption{Per-class F1 scores of BanglaBERT.}
\label{tab:per_class_banglabert}
\small
\setlength{\tabcolsep}{3pt}
\begin{minipage}[t]{0.49\linewidth}
\centering
\begin{tabular}{|l|c|}
\hline
\textbf{Rhetorical Label} & \textbf{F1} \\
\hline
Contrast & 0.590 \\
Emotional Language & 0.551 \\
Exaggeration & 0.597 \\
Metaphor & 0.656 \\
None & 0.794 \\
Repetition & 0.646 \\
Rhetorical Question & 0.752 \\
Anecdote & 0.666 \\
\hline
\end{tabular}
\end{minipage}
\hfill
\begin{minipage}[t]{0.49\linewidth}
\centering
\begin{tabular}{|l|c|}
\hline
\textbf{Persuasion Label} & \textbf{F1} \\
\hline
Blame Assignment & 0.564 \\
Call to Action & 0.681 \\
Emotional Appeal & 0.625 \\
Logical Appeal & 0.586 \\
Moral Appeal & 0.576 \\
None & 0.840 \\
Unity Call & 0.813 \\
\hline
\end{tabular}
\end{minipage}
\end{table}

\subsection{Discussion}
BanglaBERT achieves the highest performance on both tasks, suggesting that its Bangla-specific pretraining is effective for this fine-grained classification setting and that bag-of-words representations are insufficient for capturing rhetorical and persuasive cues in Bangla political speech. XLM-RoBERTa-Base performs competitively but remains below BanglaBERT, indicating that multilingual representations do not fully capture Bangla-specific rhetorical and political phrasing.

Controlled downsampling improves macro-F1 across all evaluated models for both tasks, with the largest gain observed for BanglaBERT on rhetorical detection (+3.26 points); the effect is more mixed for the remaining models, where persuasion detection sometimes benefits more, possibly because \textit{Call to Action}, \textit{Unity Call}, and \textit{None} carry clearer lexical or functional cues than the rhetorical categories.

For context, BanglaBERT reaches 88.10\% accuracy on Motamot's binary political sentiment task \cite{faria2024motamot}, a considerably coarser problem than the 9-way and 7-way distinctions modeled here. A closer comparison is SemEval-2020 Task 11, a similarly fine-grained propaganda technique classification task (14 techniques) for English news, where the best of 32 systems reached 62.1\% micro-F1 and several BERT-based systems reported macro-F1 of 0.32--0.45 under comparable imbalance \cite{dasanmartino2020semeval}. Our 65.40\% and 66.46\% macro-F1 are broadly in line with, and in some respects ahead of, this closest related benchmark, suggesting the difficulty observed here is typical of technique-level classification generally.

Confusion matrices show model errors are strongly tied to semantic overlap: for rhetorical detection, confusion among \textit{Emotional Language}, \textit{Repetition}, and \textit{Exaggeration} suggests expressive political language often combines intensity, repetition, and emotional wording, while for persuasion detection, confusion among \textit{Blame Assignment}, \textit{Moral Appeal}, and \textit{Emotional Appeal} suggests statements often combine blame, emotion, and value-based argumentation within the same segment.

\section{Limitations and Ethical Considerations}
\label{sec:limitations}

\subsection{Limitations}

Several limitations remain. First, rhetorical and persuasion annotation is inherently subjective, especially for semantically overlapping categories such as \textit{Emotional Language} vs.\ \textit{Exaggeration} and \textit{Emotional Appeal} vs.\ \textit{Moral Appeal}; the Cohen's kappa scores of 0.67 and 0.71 indicate moderate to substantial agreement, but some ambiguity remains unavoidable in fine-grained political discourse annotation. Second, despite controlled downsampling and class-weighted loss, the benchmark remains imbalanced and minority or overlapping categories are harder to classify, as reflected in the moderate macro-F1 and per-class scores. The downsampling itself was applied through manual inspection of the co-occurrence table rather than an automated, parameterized rule; we release the exact resulting counts and data for reproducibility, but future work should formalize this step. The proportion of segments requiring manual condensation was not logged during annotation and cannot be reconstructed retrospectively. Third, the data are temporally and contextually bounded: collected articles span 2014--2025 and are mainly drawn from Bangladeshi political news sources, so performance may vary across time periods, speakers, or Bangla-speaking regions. Finally, both tasks are formulated as single-label classification using the dominant label per segment, though political statements may contain multiple rhetorical or persuasive strategies, making multi-label modeling an important direction for future work.

\subsection{Ethical Considerations}

All data were collected from publicly accessible news articles and speeches. The benchmark analyzes linguistic and rhetorical patterns only, not the truthfulness or political validity of any statement, and does not endorse or rank any political individual or group. It should be used only for research and education, not political profiling or voter manipulation; all extraction and annotation were performed manually by the authors.

\section{Conclusion}
\label{sec:conclusion}

This paper presents a transformer-based benchmark for Bangla rhetorical and persuasion detection tasks using BanglaRhet, a manually annotated corpus of 30,289 political speech segments. We evaluated Bangla-specific and multilingual transformer models, including Bangla-BERT-Base, SahajBERT, BanglaBERT, and XLM-RoBERTa-Base. BanglaBERT achieved the highest performance, with 65.40\% macro-F1 for rhetorical detection and 66.46\% macro-F1 for persuasion detection. All transformer models outperformed tuned classical TF-IDF baselines by a wide margin, confirming that this task requires
representations beyond bag-of-words features.

The ablation results show that controlled downsampling improves macro-F1 across all evaluated models, most notably for BanglaBERT on rhetorical detection. A source-grouped re-evaluation further confirms these results are not substantially inflated by source-level leakage, with macro-F1 dropping by only about one point under a fully source-disjoint split. Confusion matrices and per-class analyses further show that errors are concentrated among semantically related categories, particularly where emotional, moral, exaggerative, and blame-oriented language overlap. Overall, the results establish initial benchmark baselines and indicate that analysis of rhetorical and persuasion-aware political discourse in Bangla remains challenging. Future work will focus on broader Bangla-speaking political contexts, improved minority-class modeling, source-level evaluation, and context-aware or multi-label approaches.

\bibliographystyle{IEEEtran}
\bibliography{references}

\end{document}